\documentclass[10pt,twocolumn,letterpaper]{article}

\usepackage{wacv}
\usepackage{times}
\usepackage{epsfig}
\usepackage{graphicx}
\usepackage{amsmath}
\usepackage{amssymb}

\usepackage{bbm}
\usepackage{makecell}
\usepackage{multirow}
\usepackage{graphicx}
\graphicspath{{imgs/}}

\def\wacvPaperID{488 (Applications)} 

\wacvfinalcopy 

\ifwacvfinal
\def\assignedStartPage{1} 
\fi

\ifwacvfinal
\usepackage[breaklinks=true,bookmarks=false]{hyperref}
\else
\usepackage[pagebackref=true,breaklinks=true,colorlinks,bookmarks=false]{hyperref}
\fi

\ifwacvfinal
\else
\fi

\begin{document}

\title{Scalable Temporal Localization of Sensitive Activities in Movies and TV Episodes}

\author{Xiang Hao, Jingxiang Chen, Shixing Chen, Ahmed Saad, and Raffay Hamid\\
Amazon Prime Video\\
{\tt\small \{xianghao,jxchen,shixic,hmsaad,raffay\}@amazon.com}
}
\maketitle

\begin{abstract}
	\noindent To help customers make better-informed viewing choices, video-streaming services try to moderate their content and provide more visibility into which portions of their movies and TV episodes contain age-appropriate material (\textit{e.g.}, nudity, sex, violence, or drug-use). Supervised models to localize these sensitive activities require large amounts of clip-level labeled data which is hard to obtain, while weakly-supervised models to this end usually do not offer competitive accuracy. To address this challenge, we propose a novel Coarse2Fine network designed to make use of readily obtainable video-level weak labels in conjunction with sparse clip-level labels of age-appropriate activities. Our model aggregates frame-level predictions to make video-level classifications and is therefore able to leverage sparse clip-level labels along with video-level labels. Furthermore, by performing frame-level predictions in a hierarchical manner, our approach is able to overcome the label-imbalance problem caused due to the rare-occurrence nature of age-appropriate content. We present comparative results of our approach using $41,234$ movies and TV episodes ($\sim$$3$ years of video-content) from $521$ sub-genres and $250$ countries making it by far the largest-scale empirical analysis of age-appropriate activity localization in long-form videos ever published. Our approach offers $107.2$\% relative mAP improvement (from $5.5$\% to $11.4$\%) over existing state-of-the-art activity-localization approaches.

	\vspace{-0.35cm}
\end{abstract}

\section{Introduction}
\begin{figure}[t]
    \centering
    \includegraphics[width=1.0\linewidth]{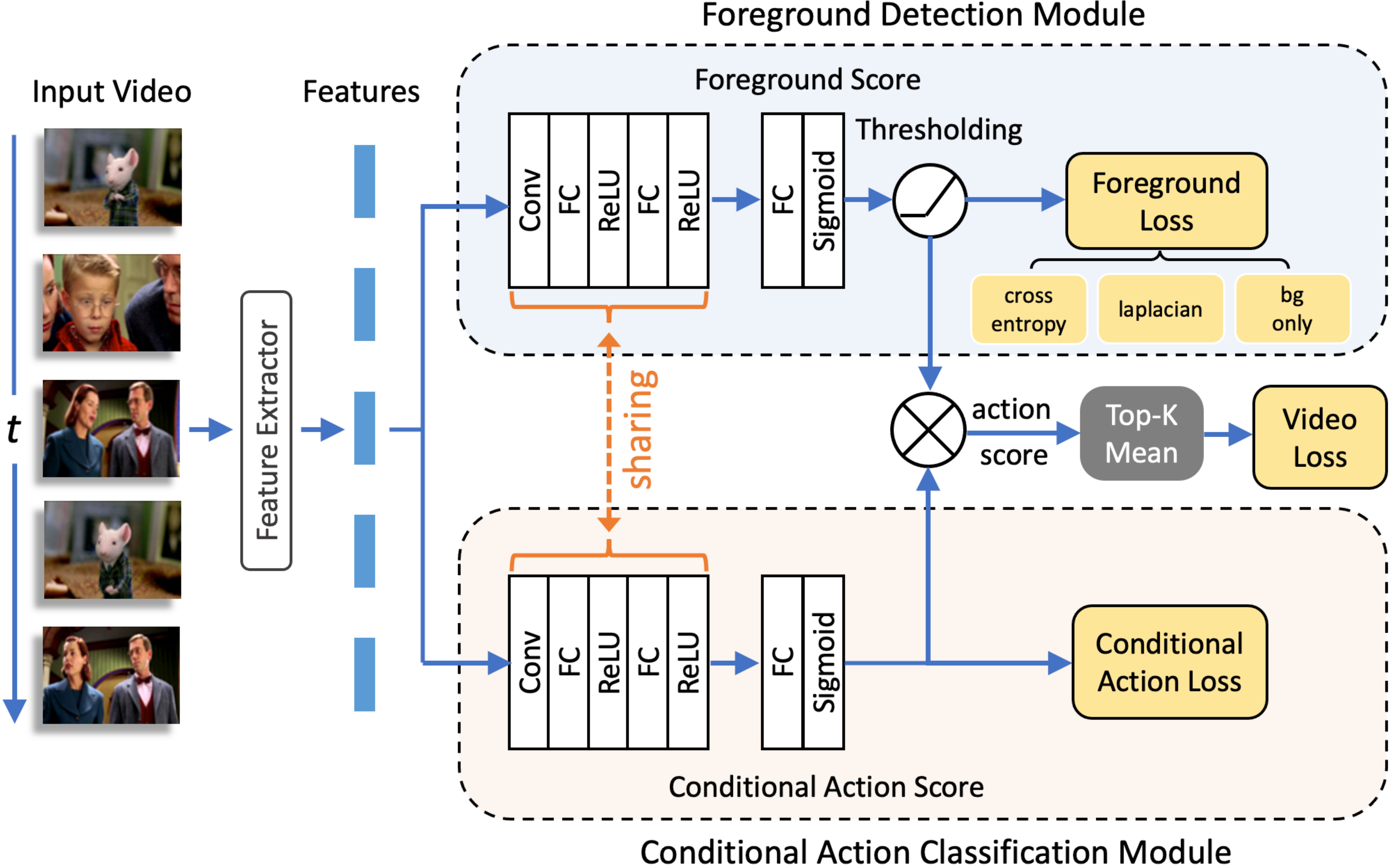}
    \caption{\textbf{Approach Overview -- } Our Corse2Fine network consists of a foreground module that predicts the binary probability of each frame belonging to the \textit{foreground} (\textit{i.e.}, containing any sensitive action), and a conditional action module which is only provided with foreground-frames using which the conditional probability of action-classes given foreground-frames is learned. The product of foreground score and action score serves as frame-level action classification score and is aggregated across the entire video using top-k mean to output a video-level classification score.}
    \label{fig:network_overview}
    \vspace{-0.2cm}
\end{figure}

\noindent Accurate and scalable content moderation has emerged as one of the most pressing challenges faced by all video-streaming services. To maintain trust with their customers and help them make more informed viewing choices, streaming services try to moderate their content and provide visibility to their customers about which parts of their videos contain age-appropriate material (\textit{e.g.}, $\mathsf{sex}$, $\mathsf{nudity}$, $\mathsf{violence}$, or $\mathsf{drug}$-$\mathsf{use}$). This visibility is particularly important for long-form videos as it can enable customers to proactively skip sensitive-content and enjoy watching movies and TV episodes confidently with their families. Our work undertakes a dedicated and thorough analysis of sensitive activity localization in movies and TV episodes to enable scalable moderation of long-form video-content.

A key challenge in building fully supervised models~\cite{buch2017end}~\cite{chao2018rethinking}~\cite{shou2016temporal}~\cite{zeng2019graph} for sensitive activity localization is the large amounts of clip-level labels required to train them which is costly and time-consuming. While weakly supervised temporal action localization methods~\cite{nguyen2019weakly}~\cite{lee2020background}~\cite{liu2019completeness}  do not pose this constraint, they are generally not as accurate as fully supervised localization approaches.

To address this challenge specifically for movies and TV episodes, it is helpful to keep in mind some of the standard practices in video-streaming industry. Most streaming services provide only video-level information about whether a movie or TV episode contains sensitive content. It is more practical for streaming-services to start with providing only video-level information indicating if an entire movie or TV episode contains any type of sensitive activity independent of its severity. This video-level binary labelling is performed by manually finding the first-occurrence of each sensitive activity-type based on which the entire video is provided a video-level label for that activity-type.

Our approach for sensitive activity localization is particularly geared to make use of these readily obtainable video-level labels in conjunction with the sparse clip-level labels based on the first-occurrence of sensitive activities. To this end, we use our proposed Coarse2Fine network (see Figure~\ref{fig:network_overview} for illustration) to make frame-level predictions about whether any sensitive activity is observable in individual frames. These frame-level predictions allow us to compare them to the clip-level labels and at the same time aggregate them at video-level to leverage the video-level labels. Our results demonstrate that this way of using weak and strong labels results in significantly more accurate activity localization compared to existing approaches.

Another key challenge in localizing age-appropriate activities in movies and TV episodes is the severe label-imbalance caused due the rare-occurrence nature of such activities ($\sim2\%$ of video-length in our data). Previous works employing background suppression or foreground modeling that indirectly addresses this challenge ~\cite{ma2020sf}~\cite{liu2019completeness}~\cite{nguyen2019weakly}~\cite{moniruzzaman2020action}~\cite{lee2020background}~\cite{lin2018bsn}~\cite{heilbron2016fast}~\cite{de2016online}~\cite{gao2017red}~\cite{baptista2019rethinking} have mainly focused on regular activities in short-form videos where the label-imbalance is not that severe (\textit{e.g.}, ActivityNet $1.3$~\cite{caba2015activitynet} and  Thumos14~\cite{THUMOS14} have $64\%$ and $29\%$ foreground frames).



To address this challenge, our Coarse2Fine network decomposes the frame-level prediction into two parts. We first perform a binary classification to identify frames containing any type of sensitive activity (called foreground-frames), followed by predicting action-classes for each of the detected foreground-frames. We compute foreground scores by introducing a novel foreground loss composed of three components: (i) a binary cross entropy loss that leverages the labels derived from the first-occurrence labels, (ii) a total-mass loss on background-only videos to minimize the foreground score on background frames, and (iii) a Laplacian regularizer that penalizes fluctuations of the foreground score. We found that our way of problem decomposition improves the label-imbalance ratio by $30$$\times$ on average, and results in localization accuracy improvement by up to $7.4\%$.


We show the effectiveness of our approach by presenting results using $41,234$ movies and TV episodes ($\sim$$3$ years of video-content) from $521$ sub-genres and $250$ countries, making it by far the largest-scale empirical analysis of age-appropriate activity localization in long-form videos ever published. Our approach is able to offer $107.2$\% relative mAP improvement (from $5.5$\% to $11.4$\%) over existing state-of-the-art activity-localization approaches.



\section{Related Work}
\noindent Our work closely relates to the areas of action recognition, and fully or weakly supervised temporal action localization. Below, we briefly go over previous approaches in these domains and distill how our approach is different from them.

\vspace{0.1cm}\noindent \textbf{a. Action Recognition:} The problem of action recognition aims to detect targeted actions using trimmed videos and has been explored from a variety of different perspectives~\cite{kong2018human}. Recent methods for action recognition include multi-stream networks~\cite{simonyan2014two}~\cite{Wang2016} that incorporate optical flow~\cite{horn1981determining} for learning motion signals and 2-D~\cite{lecun1989backpropagation} or 3-D CNNs~\cite{carreira2017quo}~\cite{tran2015learning} for learning their appearance. These networks have been shown to outperform traditional shallow-learning~\cite{ke2007spatio} or hand-crafted techniques~\cite{scovanner20073} and have demonstrated excellent performance on a variety of action recognition tasks~\cite{tran2015learning}~\cite{zhang2016real}. Besides supervised approaches, there have also been efforts to take on action recognition using self-supervised~\cite{kim2019self} and few-shot learning~\cite{zhu2020label}.

\vspace{0.1cm}\noindent \textbf{b. Fully-Supervised Temporal Action Localization:} The goal of temporal action localization (TAL) is to identify action classes in untrimmed videos as well as the start and end timestamps of each action~\cite{shou2016temporal}~\cite{chao2018rethinking}~\cite{zeng2019graph}~\cite{yeung2016end}~\cite{buch2017end}~\cite{lin2017single}. Example approaches include multi-stage CNNs \cite{shou2016temporal}, combination of CNNs and RNNs~\cite{ma2016learning}, and action models based on spatio-temporal feature representations~\cite{gkioxari2015finding}. Modern solutions of TAL use a two-stage approach where the first stage focuses on generating region proposals while the second stage focuses on action classification and boundary refinement~\cite{zhao2017temporal}~\cite{shou2016temporal}~\cite{zeng2019graph}. There also exist several end-to-end methods that combine the proposal generation and action classification steps~\cite{buch2017end}~\cite{yeung2016end}~\cite{lin2017single}. Recently, a gaussian kernel has also been proposed~\cite{long2019gaussian} to optimize temporal accuracy of action proposals. All of these methods require full annotation of start and end timestamps of each action which is expensive and time-consuming to acquire.

\vspace{0.1cm} \noindent \textbf{c. Weakly-Supervised Temporal Action Localization:} The goal of weakly-supervised temporal action localization (W-TAL) is to require less supervision by only relying on video-level labels. For example, work in~\cite{paul2018w} formulates W-TAL as a multi-instance learning problem while~\cite{nguyen2018weakly} employs class-agnostic attention with sparsity regularization. Similarly, work in~\cite{liu2019completeness} propose a multi-branch network to localize actions more accurately and proposes a scheme for generating background videos for better context separation. This work inspired subsequent works~\cite{nguyen2019weakly}~\cite{lee2020background} that try to model the background class for more accurate temporal localization. The performances of W-TAL methods is generally still not at par with fully-supervised TAL methods.

Recently, SF-Net~\cite{ma2020sf} has explored a middle ground between TAL and W-TAL by using supervision from a single frame of each action-occurrence in addition to the video-level labels for action localization. Similar to SF-Net~\cite{ma2020sf}, our work aims to find an intermediate form of supervision; however, with two key differences. First, unlike requiring supervision from a single-frame of\textbf{ every single} action instance. In contrast, we use the annotation of \textbf{only the first-occurrence} of an action. Second, our work focuses on long-form videos which contain significantly high number of background frames which makes the technical challenge of our problem significantly greater than that taken on by SF-Net~\cite{ma2020sf}. Note that most previous action localization approaches show competitive results only on short-form videos~\cite{THUMOS14}~\cite{caba2015activitynet} where durations of the majority of videos are up to a few minutes. So far, only limited success has been reported on relatively long videos~\cite{gu2018ava}~\cite{girdhar2018better}.


\section{Method}
\subsection{Problem Formation}
\noindent Let $\textrm{C}$ denote different action classes and $\textrm{N}$ represent the training videos $\{v_n\}_{n=1}^\textrm{N}$ with video-level labels $\{y_n\}_{n=1}^\textrm{N}$. Here $y_n$ is a $\textrm{C}$-dimensional binary vector and $y_{n;c}$ is $1$ if the $n$-th video contains $c$-th action class. 
Unlike the fully-supervised setting which provides the start and end timestamps of all action occurrences~\cite{buch2017end}~\cite{chao2018rethinking}~\cite{zeng2019graph}, in our setting only the first occurrence of each action is labeled with known start and end timestamps $(s, t)$. Additionally, a labeled action instance can contain multiple actions \textit{e.g.}, $\mathsf{sex}$ and $\mathsf{nudity}$ occurring simultaneously, which makes ours a multi-label classification problem.

Our proposed network-architecture is shown in Figure~\ref{fig:network_overview}. During training, videos are provided to our network to generate two frame-level scores: (i) a binary classification score predicting the probability of each frame being foreground \textit{i.e.}, containing any action of interest, and (ii) a $\textrm{C}$-dimensional conditional action classification score indicating the probability of each action given that there exists an action in the frame. The foreground score is optimized using a foreground loss comprising of three components: (i) a binary cross entropy loss that leverages the labels derived from the first-occurrence labels, (ii) a total-mass loss on background-only videos to minimize the foreground score on background frames, and (iii) a Laplacian regularizer that penalizes fluctuations of the foreground score. 

Our conditional action module is trained with multi-label classification loss with labels derived from the first-occurrence data and is only provided with frames where there is at least one action to force the module to learn a frame's conditional probability of action-class given that it is a foreground frame. The product of foreground score and conditional action score serves as frame-level action score which is aggregated over the video to produce a video-level classification score thereon provided to a video-level multi-label classification loss function for training. 

During inference, action intervals are proposed by thresholding the frame-wise action score. In the following sections, we discuss the details of each module of our network structure.

\subsection{Coarse2Fine Network}

\subsubsection{Feature Extraction}
For a training batch of $\textrm{N}$ videos, a pre-trained feature extractor (see Section~\ref{sec:experiment_setup} for details) is used to generate $\textrm{D}$-dimensional vectors extracted from all frames, and stored in a feature tensor $\textrm{X} \in \textrm{R}^{\textrm{N}\times\textrm{T}\times\textrm{D}}$, where $\textrm{T}$ is the number of frames. We perform zero-padding when the number of frames in a video is less than $\textrm{T}$. 

\subsubsection{Addressing Label Imbalance} 
The rare-occurrence nature of sensitive activities in movies and TV episodes ($\sim$$2$\% of video-length in our data) results in there being label-imbalance for all the action-classes. To address this challenge, we propose to decompose the prediction of frame-level action classification into two sub-problems. First, we find foreground frames from all frames in a video that contain any action of interest. Second, we identify actions that exist in the detected foreground frames.

This problem decomposition can be formulated in terms of a probabilistic graphical model~\cite{koller2009probabilistic} as following:
\begin{equation}
\textrm{P}(\textrm{A} | \textrm{X}) = \textrm{P}(\textrm{F} = \textrm{1} | \textrm{X}) \cdot \textrm{P}( \textrm{A} | \textrm{F} = \textrm{1},\textrm{X})
\end{equation}

\noindent where $\textrm{A}$ denotes action variable, $\textrm{F}$ denotes binary foreground variable, and $\textrm{X}$ denotes frame features. This way of problem decomposition improves the label-imbalance ratio by $30$$\times$ on average, and results in localization accuracy improvement by up to $7.4\%$ (see $\S$~\ref{ss:data_deep_dive} and~\ref{ss:results} for details).



\subsubsection{Foreground Detection Module}
The foreground detection module outputs a score indicating the probability of presence of any sensitive action for each frame in a video. To classify each frame, we provide $\textrm{X}_n$ to temporal $1$-$\textrm{D}$ convolutional (CONV) and fully-connected (FC) layers to get the foreground score $\textrm{F}_n \in \textrm{R}^{\textrm{N}\times\textrm{T}}$ after sigmoid function. This can be formalized as follows: 
\begin{equation}
\textrm{P}(\textrm{F}_n,\textrm{X}_n) = \textrm{sigmoid}(f_{\textrm{FC}}(f_{\textrm{CONV}}(\textrm{X}_n; \theta_c), \theta_f))
\label{eq:fore_det}
\end{equation}

\noindent where $\theta_c$ and $\theta_f$ denote trainable parameters in the convolutional and FC layers respectively. The foreground score $\textrm{F}_n$ is first thresholded to zero-out all elements below a threshold $\epsilon$ and encourage the model to detect only highly discriminative foreground frames. This thresholded foreground score is used in a binary cross entropy loss $\textrm{L}_{ce}$ with labels derived from first-occurrence clip labels for training.

To further leverage the background information in videos without any sensitive actions, we employ an additional total-mass loss on the foreground score $\textrm{P}(\textrm{F} = \textrm{1}|\textrm{X})$ from videos without any actions to minimize its $\textrm{L}_\textrm{1}$ norm, \textit{i,e.}:
\begin{equation}
\textrm{L}_{\textrm{bg-only}} = \sum_{n=1}^\textrm{N} \mathbbm{1}_n \sum_{t=1}^\textrm{T} |\textrm{P}(\textrm{F}_n^t=1|\textrm{X})|
 \end{equation}
  
\noindent where $\mathbbm{1}_n$ is an indicator function indicating if the video $v_n$ does not contain any sensitive actions.

Lastly, as we do not expect the foreground and background to change too frequently, we apply a Laplacian regularizer, $\textrm{L}_{\textrm{laplacian}} = |\nabla \textrm{P}(\textrm{F}=\textrm{1}|\textrm{X})|$ to penalize fluctuations of the foreground score $\textrm{P}(\textrm{F}=\textrm{1}|\textrm{X})$. 

Overall, our foreground loss function therefore comprises of three components: (i) a binary cross entropy loss that leverages labels derived from the first-occurrence label, (ii) a total-mass loss on background-only videos to minimize the foreground score on background frames, and (iii) a Laplacian regularization component that penalizes fluctuations of the foreground score.
\begin{equation}
\textrm{L}_{\textrm{foreground}} = \textrm{L}_{ce} + \gamma \cdot \textrm{L}_{\textrm{bg-only}} + \delta \cdot \textrm{L}_{\textrm{laplacian}}
\end{equation}

\subsubsection{Conditional Action Classification Module}
The classification module of our network outputs the conditional probability $\textrm{P}(\textrm{A}|\textrm{F}=\textrm{1},\textrm{X})$ of action classes for all frames in input videos. Similar to the foreground detection module, we provide $\textrm{X}$ to temporal $1$-D convolutional and fully-connected (FC) layers with sigmoid activation function to get the conditional action score $\textrm{P}(\textrm{A}|\textrm{F}=1, \textrm{X}) \in \textrm{R}^{\textrm{N}\times\textrm{T}\times\textrm{C}}$ to compute the classification loss. The convolutional and FC layers (except for the last FC layer) are shared with foreground detection module. During training, we only provide frame samples where there is at least one action in order to force the module to learn a frame's conditional probability of action class given that it is known to be a foreground frame, and the classification score is used to compute frame's multi-label classification loss $\textrm{L}_{\textrm{cs}}$ with labels derived from the first-occurrence clip labels.

\subsubsection{Title-Score Aggregation}
Before we aggregate the computed frame-level scores into a single video-level class score for comparing to the video-level ground truth, we first multiply the foreground score $\textrm{P}(\textrm{F}=\textrm{1}|\textrm{X})$ with the conditional action score $\textrm{P}(\textrm{A}|\textrm{F}=\textrm{1},\textrm{X})$ to get actual action score $\textrm{P}(\textrm{A}|\textrm{X}) \in \textrm{R}^{\textrm{N}\times\textrm{T}\times\textrm{C}}$. We then adopt the top-k mean technique~\cite{paul2018w}~\cite{wang2017untrimmednets} on the frame-level action score. The video-level class-score $y_n^c$ for class $c \in \textrm{C}$ of video $v_n$ can be derived as follows:
\begin{equation}
y_n^c = \frac 1 k \cdot \max_{S \subset \textrm{P}(\textrm{A}_n^c|\textrm{X}), |\textrm{S}| = k} \sum_{\forall s \in \textrm{S}}  s
\end{equation}

\noindent which is used for video-level cross entropy loss $\textrm{L}_{\textrm{video}}$.

Our model is trained jointly with all three losses, and the overall loss function is composed as follows:

\begin{equation*}
\textrm{L} = \textrm{L}_{\textrm{video}} + \alpha \cdot \textrm{L}_{\textrm{foreground}} + \beta \cdot \textrm{L}_{\textrm{cs}},
\end{equation*}
where $\alpha$, $\beta$, $\gamma$, and $\delta$ are the hyper-parameters. 

\subsection{Inference}
\noindent During inference, following the work of~\cite{ma2020sf, lee2020background}, we discard classes whose video-level probabilities are below the pre-defined threshold. For the remaining categories, we threshold the action score $\textrm{P}(\textrm{A}|\textrm{X})$ to select candidate segments and compute the segments' scores for each proposal using the highest scores among the detected segments.

\section{Experiments}
\subsection{Data Deep-Dive}
\label{ss:data_deep_dive}
\noindent Most existing action data-sets (\textit{e.g.},
Thumos14~\cite{THUMOS14}, or ActivityNet $1.3$~\cite{caba2015activitynet})
mainly focus on everyday actions in short-form videos. At the same time, the few
long-form video data-sets (\textit{e.g.} AVA~\cite{gu2018ava}, MovieNet~\cite{huang2020movienet}) do not focus on detecting and localizing age-appropriate activities.

To undertake a thorough and focused analysis of our problem, we therefore introduce a new data-set containing $41,234$ videos with $86$\% TV episodes and $14$\% movies. These videos have a mean duration of $38$ minutes ($2.97$ years of total content), span $521$ sub-genres, and are from $250$ countries making it the largest-scale data-set for age-appropriate activity localization in long-form videos.

We focus on four types of sensitive activities, \textit{i.e.},: (a) $\mathsf{sex}$, (b) $\mathsf{nudity}$, (c) $\mathsf{violence}$, and (d) $\mathsf{drug}$-$\mathsf{use}$. To have consistent and reliable annotations among our $34$ annotators, we provided annotation guidelines containing: $17$ instructions for $\mathsf{sex}$, $11$ for $\mathsf{nudity}$, $34$ for $\mathsf{violence}$ and $14$ for $\mathsf{drug}$-$\mathsf{use}$. Examples of these instructions include: ``intentional murder and/or suicide that involves bloody injury" for $\mathsf{violence}$, and
``fully exposed buttocks in a sexual context" for $\mathsf{nudity}$.  All $34$ annotators were fluent English speakers and went through multiple training sessions to prepare for their assignment. To ensure that the annotators had been properly trained, we asked each annotator to label sampled videos and received feedback on their labeling quality before they could start to label the data independently.

\begin{figure*}[t]
    \centering
    \includegraphics[width=1.0\linewidth]{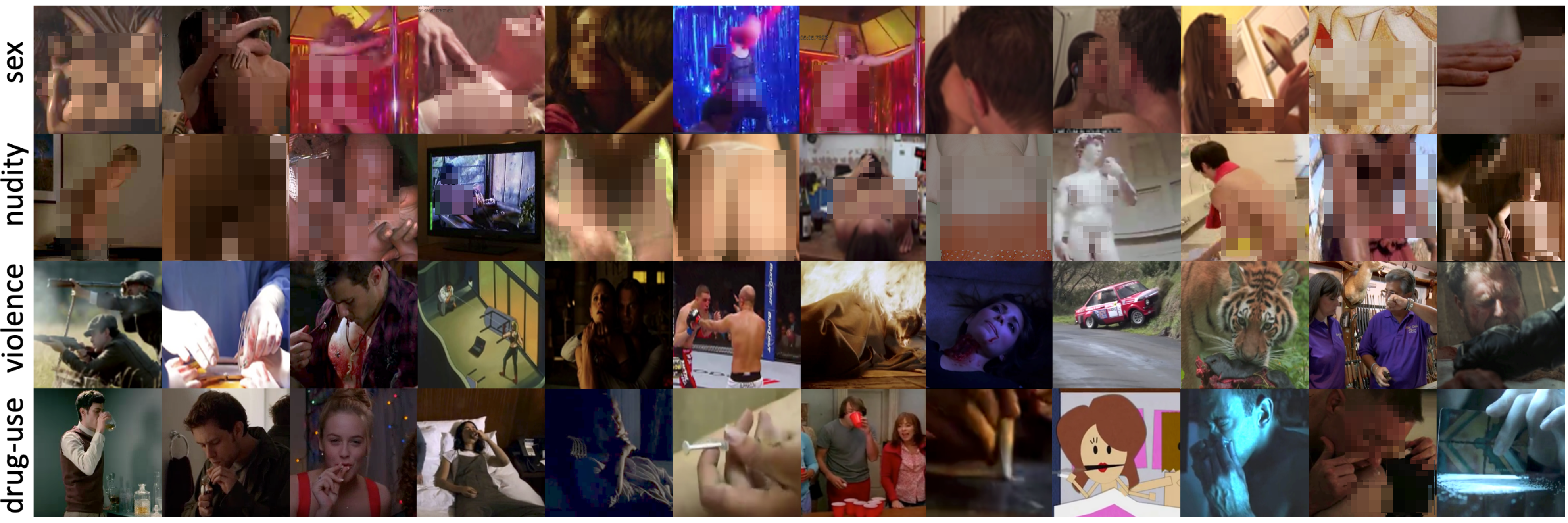}
    \caption{Examples of the four types of age-appropriate activities considered
      in our data-set. We have redacted sensitive parts
      of the images on purpose to make them non-offensive. Primarily,
      $\mathsf{sex}$ includes sexual activities, sexual touching, strip-tease,
      erotic dancing, and kissing. $\mathsf{Nudity}$ includes display of
      genitalia, adult buttocks, and female breasts. $\mathsf{Violence}$
      includes injuries or death caused by violence, sexual violence, depiction
      of blood, fight or weapons, and sports injury (such as race car crash and
      boxing). $\mathsf{Drug}$-$\mathsf{use}$ includes consumption, depiction,
      or appearance of illegal drugs, tobacco and alcohol. }
    \label{fig:example}
\end{figure*}

During labeling, each video was played at $24$ frames per second, and the annotators checked the video from the beginning for a list of elements related to the four types of age-appropriate activities. The corresponding start and end timestamps of the first-occurrence of each activity were recorded in case the activity existed, and were left blank otherwise. To ensure label quality, a random sample of labels was selected and taken through a mandatory review.

Each of the four types of age-appropriate activities in our data contains a wide
range of variation. Overall, $\mathsf{sex}$ primarily includes sexual activity
(including body action and facial expressions), sexual touching, strip-tease,
erotic dancing, and kissing. Nudity includes display of genitalia, adult
buttocks, and female breasts. Violence mainly includes injuries or death caused
by violence (such as murder and suicide), sexual violence (\textit{e.g.}, rape),
depiction of blood, fight (including animals such as tiger hunting) or weapons
(\textit{e.g.}, guns and knives), and sports injury (such as race car crash and
boxing). $\mathsf{Drug}$-$\mathsf{use}$ primarily includes consumption,
injection, depiction, or description on illegal drug-products (such as heroin
and marijuana), tobacco and various alcohol usage. Some of the representative
examples of these actions are provided in Figure~\ref{fig:example}. Detecting
those content can be challenging even for human beings considering the numerous
variations of those actions that leads to a long tail distribution of the related
objects, \textit{i.e.}, over 100 different objects are closely related to these four types of
actions. Additionally, certain categories may depend on significant context to
be identified such as fight and alcohol usage.

\begin{table}[t]
  \begin{center}
    \begin{tabular}{c|r|r|r|r}
      \hline
          & \makecell{$\mathsf{sex}$} & \makecell{$\mathsf{nudity}$} &
                                                                  \makecell{$\mathsf{violence}$} & \makecell{$\mathsf{drug}$-$\mathsf{use}$}\\
      \hline
      Label-Counts      & 12,241 & 2,164 & 20,592 & 14,578 \\
      \hline
      Avg. Duration      & 23.2s & 18.9s & 28.4s & 19.3s \\
      \hline
    \end{tabular}
  \end{center}
  \caption{Age-appropriate clip label counts with average durations. As an
    example, there are $12,241$ videos containing
    $\mathsf{sex}$. For each of these $12,241$ videos, the first occurrence of
  $\mathsf{sex}$ action is manually labeled and has an average duration is $23.2$ seconds.}
  \label{tab:label_stats}
  \vspace{-0.5cm}
\end{table}

From our $41,234$ videos, $26,726$ contain at least one type of age-appropriate activity, while the remaining $14,508$ videos do not contain any of the four types of age-appropriate content. Table~\ref{tab:label_stats} shows the number of videos containing each of the activities among the $26,726$ videos. We can see that $\mathsf{violence}$ occurs more often while $\mathsf{nudity}$ occurs rarely with an average duration around $20$ seconds.

The $1$-st row of Table~\ref{tab:perc_presence} shows the estimated percentage of presence for each age-appropriate activity in our entire data-set. It can be seen that the activity-labels are quite imbalanced especially for $\mathsf{nudity}$, for which we have
only $0.2\%$ presence in all videos. In total, $2.5\%$ of all frames are
foreground frames, \textit{i.e.}, containing any type of sensitive
activity. Note that row-$1$ does not sum to $2.5\%$ because the same part of a
video can contain multiple sensitive activities. The $2$-nd row of
Table~\ref{tab:perc_presence} shows the percentage of presence for each activity
among the foreground frames only. For example, $24.8\%$ of the foreground frames
contain $\mathsf{sex}$. This presence-percentage is increased by $30$$\times$ on
average when we restrict our focus to only foreground frames. 



\begin{table}[t]
  \begin{center}
    \begin{tabular}{c|r|r|r|r}
      \hline
           \makecell{Frames} & \makecell{$\mathsf{sex}$} & \makecell{$\mathsf{nudity}$} &
                                                                  \makecell{$\mathsf{violence}$} & \makecell{$\mathsf{drug}$-$\mathsf{use}$}\\
      \hline 
      All & 0.8\% & 0.2\% & 1.3\% & 0.7\%\\
      \hline
      Foreground & 24.8\% & 3.4\% & 48.5\% & 26.0\%\\
      \hline
    \end{tabular}
  \end{center}
  \caption{The $1$-st row shows the percentage of presence for each age-appropriate activity in our entire data-set. The $2$-nd row shows the percentage of presence for each activity among the foreground frames only (\textit{i.e.}, only the frames containing any sensitive activity). Restricting focus from all frames to only foreground frames improves our label-imbalance by $30$$\times$ on average (\textit{i.e.}, $30$$\times$ for $\mathsf{sex}$, $16$$\times$ for $\mathsf{nudity}$, $36$$\times$ for $\mathsf{violence}$, and $36$$\times$ for $\mathsf{drug}$-$\mathsf{use}$).}
  \label{tab:perc_presence}
  \vspace{-0.5cm}
\end{table}


For model training, we randomly split our data-set into training ($35,818$ videos) and test
($5,416$ videos) sets. When measuring our model performance on the test data-set, since
only the first occurrence of each age-appropriate activity is labeled, we measure our model's localization performance up to the end timestamp of each activity if it exists. Otherwise, we measure it till the end of the video. We refer to this test data as \textbf{first occurrence} data. To compensate for the fact that only first-occurrence labels are available in the test-set for evaluation, we also labeled all occurrences of age-appropriate activities on a set of $28$ videos with a total duration of $1,075$ minutes. We refer to this data as \textbf{all-occurrence} data-set and use it for additional evaluation.






\subsection{Experimental Settings}
\label{sec:experiment_setup}

\vspace{0.1cm}\noindent \textbf{Features:} For our experiments, we used the ResNet50 deep network~\cite{he2016identity}
pre-trained on ImageNet~\cite{russakovsky2015imagenet} to extract frame-level visual
features of our data-set. We extracted the video-frames at $1$ frame per second
(FPS), provide them to the ResNet50 network, and then extract the last
hidden layer ($2048$ dimensions) as the frame feature. These features are provided as input to our proposed Coarse2Fine network.

\vspace{0.1cm}\noindent \textbf{Evaluation Metrics:} We follow the standard
mean average precision (mAP) metric and use the evaluation code from
ActivityNet\footnote{https://github.com/activitynet/ActivityNet/} work to measure models' performance for different values of interactions of union (IoU) thresholds.

\vspace{0.1cm}\noindent \textbf{Implementation Details:} We extract labels for
training for foreground detection and conditional action classification modules
based on the first-occurrence labels. Specifically, for foreground detection
module, our label is binary and we treat the frames inside the labeled occurrence intervals as foreground frames. The frames before the start
timestamp are candidates for background frames since the same type of sensitive activities may present after the first occurrence. We randomly select $20\%$ of
background-frame candidates for training. The conditional action classification module is trained by only using the
foreground frames. The label for each foreground frame is a
$\textrm{C}$-dimensional binary vector whose dimensions set to $1$ if the
corresponding activity-class occurs in the frame. We use $1$ CONV layer and $2$ FC layers (with 1024
and 512 neurons respectively and each followed by ReLu activation) in Equation~\ref{eq:fore_det} and fix number of
input frames $\textrm{T}$ to $3,600$. We performed parameter optimization using
Adam~\cite{kingma2014adam} with batch size $16$ and learning rate
$1e^{-5}$ and selected hyper-parameters using grid search with
$\alpha = 0.5$, $\beta=1$, $\gamma=0.01$, $\delta=0.1$, and $\epsilon=0.05$.


\subsection{Results}
\label{ss:results}
\noindent \textbf{Comparisons with State-of-the-art Methods:} We compared our model with: (i) a variant of~\cite{ma2020sf} (which we call SF-NET-$1$) that also explores a middle ground between TAL and W-TAL similar to our work, and (ii) a weakly supervised method~\cite{narayan20193c}. We only considered the appearance branch (\textit{i.e.}, the RGB stream) in all model-comparisons to remove the variance of features used in different approaches and focus on the comparison between model architectures.
When implementing~\cite{ma2020sf} for SF-NET-$1$, we picked the middle point
at each labeled clip, and used it as the frame label for the corresponding action. Additionally, as the average number of activity instances in our data-set (\textit{i.e.}, about $2$ occurrences per video) is smaller than in public
data-sets such as THUMOS14~\cite{THUMOS14} ($15$ on average), it can negatively affect
the performance of~\cite{ma2020sf} on our data-set. We therefore implemented an amplified
version of~\cite{ma2020sf} (we call SF-NET-$2$) by uniformly sampling $k$-frames frames from each action-segment. We set $k$ equal to $7$. The comparative mAP with IoUs from $0.0$ to $0.2$ are given in Table~\ref{tab:results_internal_dataset}. 

We can see that SF-Net-$1$ and $2$ perform better than $3$C-Net, while SF-Net-$2$ performs
better than SF-Net-$1$ which uses less information. Our model offers the best performance compared to all other methods on both first-occurrence as well as all-occurrence data-sets. Additionally, the alternative methods all performed poorly to localize
the actions when IoU is relative high. Similar to what was observed
in~\cite{liu2020multi} where the state-of-the-art temporal action localization
methods achieve only modest mAP in more challenging multi-shot scenario, the mAP
achieved by all methods considered in our work is relatively low compared to the
metrics reported on Thumos14~\cite{THUMOS14} and ActivityNet-$1.3$~\cite{caba2015activitynet}. This demonstrates that action localization in long-form videos is a much more challenging problem.

\begin{table}[t]
  \begin{center}
    \begin{tabular}{c|l|l|l|l|l}
      \hline

      \multirow{2}{*} {\shortstack{Data-set}} &
      \multirow{2}{*}{\shortstack {Method} } 
      & \multicolumn{4}{c} {mAP@IoU} \\\cline{3-6}
      & & 0.0 &0.1 & 0.2 & AVG\\
      \hline
      \hline
      \multirow{4} {*} {\shortstack{\makecell{First \\Occurence}}} 
      & $3$C-Net~\cite{narayan20193c}  & 8.3 & 3.7 & 1.1 & 4.4\\\cline{2-6}
      & SF-Net-$1$~\cite{ma2020sf} & 9.6 & 4.9 & 1.1 & 5.2 \\\cline{2-6}
      & SF-Net-$2$~\cite{ma2020sf} & 10.1 & 5.2 & 1.2 & 5.5 \\\cline{2-6}
      & Ours & 22.5 & 6.7 & 5.0 & 11.4\\
      \Xhline{3\arrayrulewidth}
      \multirow{4} {*} {\makecell{All \\Occurence}} 
      & 3C-Net~\cite{narayan20193c} & 23.7 & 3.6 & 0.8 &  9.4\\\cline{2-6}
      & SF-Net-$1$~\cite{ma2020sf} & 24.7 & 5.2 & 1.6 & 10.5\\\cline{2-6}
      & SF-Net-$2$~\cite{ma2020sf} & 29.8 & 6.1 & 2.1 & 12.7 \\\cline{2-6}
      & Ours & 44.9 & 19.5 & 15.9 & 26.8 \\
      \hline
    \end{tabular}
  \end{center}
  \vspace{-0.25cm}
  \caption{Performance comparison with existing state-of-the-art
    activity-localization approaches.}
  \label{tab:results_internal_dataset}
	  \vspace{-0.5cm}
\end{table}



\begin{figure*}[t]
    \centering
    \includegraphics[width=1\linewidth]{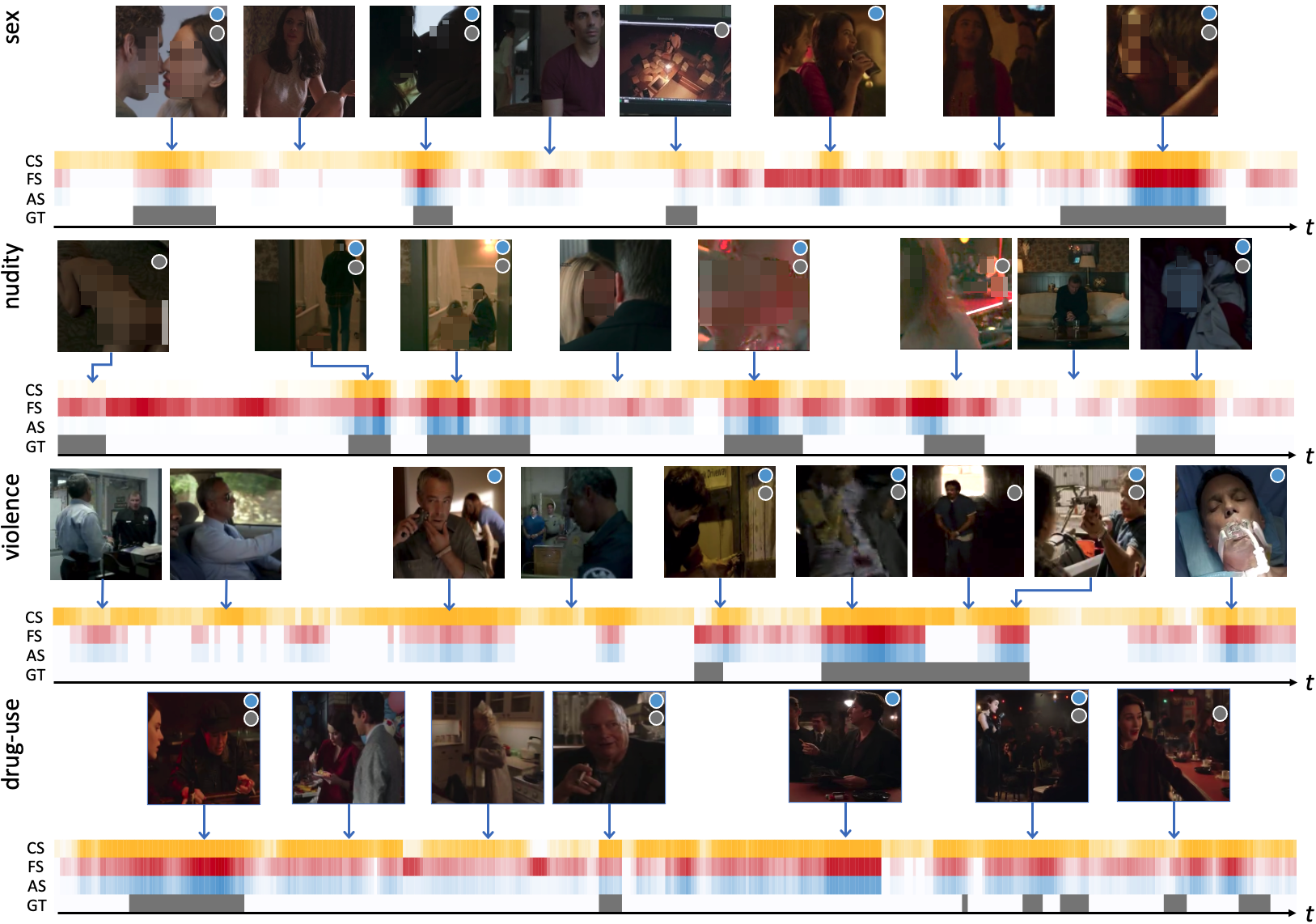}
    \caption{Visualization of the detection results for qualitative
      analysis. We have redacted sensitive parts of the
      images on purpose to make them non-offensive. The blue dot on an image
      indicates the frame was detected by the model, and the grey dot indicates
      the image was between the labeled start and end timestamps.  `CS' represents
      the conditional action score, `FS' represents the foreground score, `AS'
      represents the action score, and `GT' represents the ground truth.}
    \label{fig:qua}
      \vspace{-0.4cm}
\end{figure*}

\vspace{0.1cm}\noindent \textbf {Performance on Each Activity-Type:} We analyze our model's performance on each age-appropriate action to get more insights on the
detection accuracy of our model. The mAP metrics for each action with IoUs ranging from 0.0 to 0.2 are summarized in
Table~\ref{tab:results_each_action_internal_dataset}. We can see that our model
performs the best for $\mathsf{violence}$ action on both first-occurrence and all-occurrence data-sets followed by $\mathsf{nudity}$ and $\mathsf{sex}$ actions. For drug
use, our model performs worst compared to the other three activity-classes
as $\mathsf{drug}$-$\mathsf{use}$ often depends on long-range context and involves small objects,
such as tobacco and syringe needles, which are very challenging especially with dark background.

\begin{table}[t]
  \begin{center}
    \begin{tabular}{c|l|l|l|l|l}
      \hline

      \multirow{2}{*} {\shortstack{Data-set}} &
      \multirow{2}{*}{\shortstack {Action} } 
      & \multicolumn{4}{c} {mAP@IoU} \\\cline{3-6}
      & & 0.0 &0.1 & 0.2 & AVG\\
      \hline
      \hline
      \multirow{4} {*} {\shortstack{\makecell{First \\Occurence}}} 
      & $\mathsf{sex}$  & 21.1 & 5.5 & 4.2 & 10.3\\\cline{2-6}
      & $\mathsf{nudity}$ & 29.2 & 7.4 & 5.1 & 13.9 \\\cline{2-6}
      & $\mathsf{violence}$ & 24.1 & 10.2 & 8.1 & 14.1 \\\cline{2-6}
      & $\mathsf{drug}$-$\mathsf{use}$ & 15.7 & 3.7 & 2.6 & 7.3\\
      \Xhline{3\arrayrulewidth}
      \multirow{4} {*} {\makecell{All \\Occurence}} 
      & $\mathsf{sex}$ & 35.8 & 27.2 & 25.8 &  29.6\\\cline{2-6}
      & $\mathsf{nudity}$ & 47.0 & 19.1 & 17.1 & 27.7\\\cline{2-6}
      & $\mathsf{violence}$ & 84.8 & 29.7 & 18.7 & 44.4 \\\cline{2-6}
      & $\mathsf{drug}$-$\mathsf{use}$ & 12.0 & 2.2 & 2.0 & 5.4 \\
      \hline
    \end{tabular}
  \end{center}
  \vspace{-0.25cm}
  \caption{Performance of our model for different activity-classes.}
  \label{tab:results_each_action_internal_dataset}
    \vspace{-0.5cm}
\end{table}

\vspace{0.1cm}\noindent \textbf{Ablation Study:}
To analyze the effectiveness of the foreground module, conditional action module,
and our approach of problem decomposition, we perform a set of ablation studies on
first-occurrence and all-occurrence data-sets. The localization mAP at different
thresholds is presented in Table~\ref{tab:ablation_study}. We can see that the
model trained with only first-occurrence labels (TAL setting but with only the first occurrence) performs poorly with $3.2$ average mAP on the first-occurrence data-set. When combining the first-occurrence labels with video-level labels
without problem decomposition, the performance is improved to $4.0$ average mAP. Lastly, using our approach for problem decomposition, along with video-level and first-occurrence labels, our Coarse2Fine network is able to achieve $11.4$ average mAP.

\begin{table}[t]
  \begin{center}
    \begin{tabular}{c|l|l|l|l|l}
      \hline

      \multirow{2}{*} {\shortstack{Data-set}} &
                                                                     \multirow{2}{*}{\shortstack {Method} } 
      & \multicolumn{4}{c} {mAP@IoU} \\\cline{3-6}
      & & 0.0 & 0.1 & 0.2 & AVG\\
      \hline
      \hline
      \multirow{4} {*} {\shortstack{\makecell{First \\Occurence}}} 
      & FO & 6.7 & 1.7 &1.2  & 3.2\\\cline{2-6}
      & + VL & 7.9 & 2.4 & 1.6  & 4.0 \\\cline{2-6}
      & + VL + PD & 22.5 & 6.7 & 5.0  & 11.4\\
      \Xhline{3\arrayrulewidth}
      \multirow{4} {*} {\makecell{All \\ Occurence}} 
      & FO & 26.8 & 6.7 & 5.4 & 13.0\\\cline{2-6}
      & + VL &31.8  & 10.0 & 6.8  & 16.2\\\cline{2-6}
      & + VL + PD & 44.9 & 19.5  & 15.9 & 26.8\\
      \hline
    \end{tabular}
  \end{center}
  \vspace{-0.25cm}
  \caption{Ablation study: FO uses only first occurrence labels for frame level
    supervision when training. + VL indicates adding video level
  supervision, and + PD means the model is using the problem decomposition in our
  Coarse2Fine network.}
  \label{tab:ablation_study}
  \vspace{-0.5cm}
\end{table}

\subsection{Qualitative Results}
\noindent Qualitative examples of our localization result along with ground-truth labels are shown in
Figure~\ref{fig:qua}. Each example represents one
of the age-appropriate activity-types ($\mathsf{sex}$, $\mathsf{nudity}$,
$\mathsf{violence}$ and $\mathsf{drug}$-$\mathsf{use}$). We use heat-maps to
represent model prediction and ground-truth. The grey bar denotes ground-truth (GT), red denotes estimated foreground
scores (FS), yellow denotes conditional action scores(CS), and blue
denotes the action scores (AS) adjusted by the foreground scores. The presence of a blue
dot on an image indicates if the frame was detected by the model, while the grey dot
indicates if the image was between the start and end timestamps of labeled sensitive activity.

For action-type $\mathsf{sex}$ our model detects kissing and sexual
actions in the $1$-st, $3$-rd and last images, but misses it in cases where for
instance the activity occupies a small portion of the frame, as shown in $5$-th image.


For action-type $\mathsf{nudity}$, the model is generally able to make correct
predictions and localize the locations where $\mathsf{nudity}$ is
present. Examples include naked female body and buttocks. We also present one example of false negative ($6$-th image) where erotic dancing in the background behind the main character is happening.

For action-type $\mathsf{violence}$, the positive examples include gunshot,
getting injured, and break-in, with the key challenge being the variance of the backgrounds where these actions happen. Overall, our model is able to localize $\mathsf{violence}$ well except for clips with dark illumination (image-$7$).

In the $\mathsf{drug}$-$\mathsf{use}$ example, the drug related behaviors
include smoking (cigarette and cigar), and alcohol usage in a banquet. The
model succeeded in capturing most of the positive actions except when there is a human drinking in the background and the camera does
not zoom in enough, as shown in the last image. Also, foreground scores help correct
several detections from the conditional action scores (yellow bar), \textit{e.g.}, eating at a social event ($2$-nd image) and picking a cup in a kitchen ($3$-rd image). 

\subsection{Error Analysis}
\begin{figure}[t]
    \centering
    \includegraphics[width=1.0\linewidth]{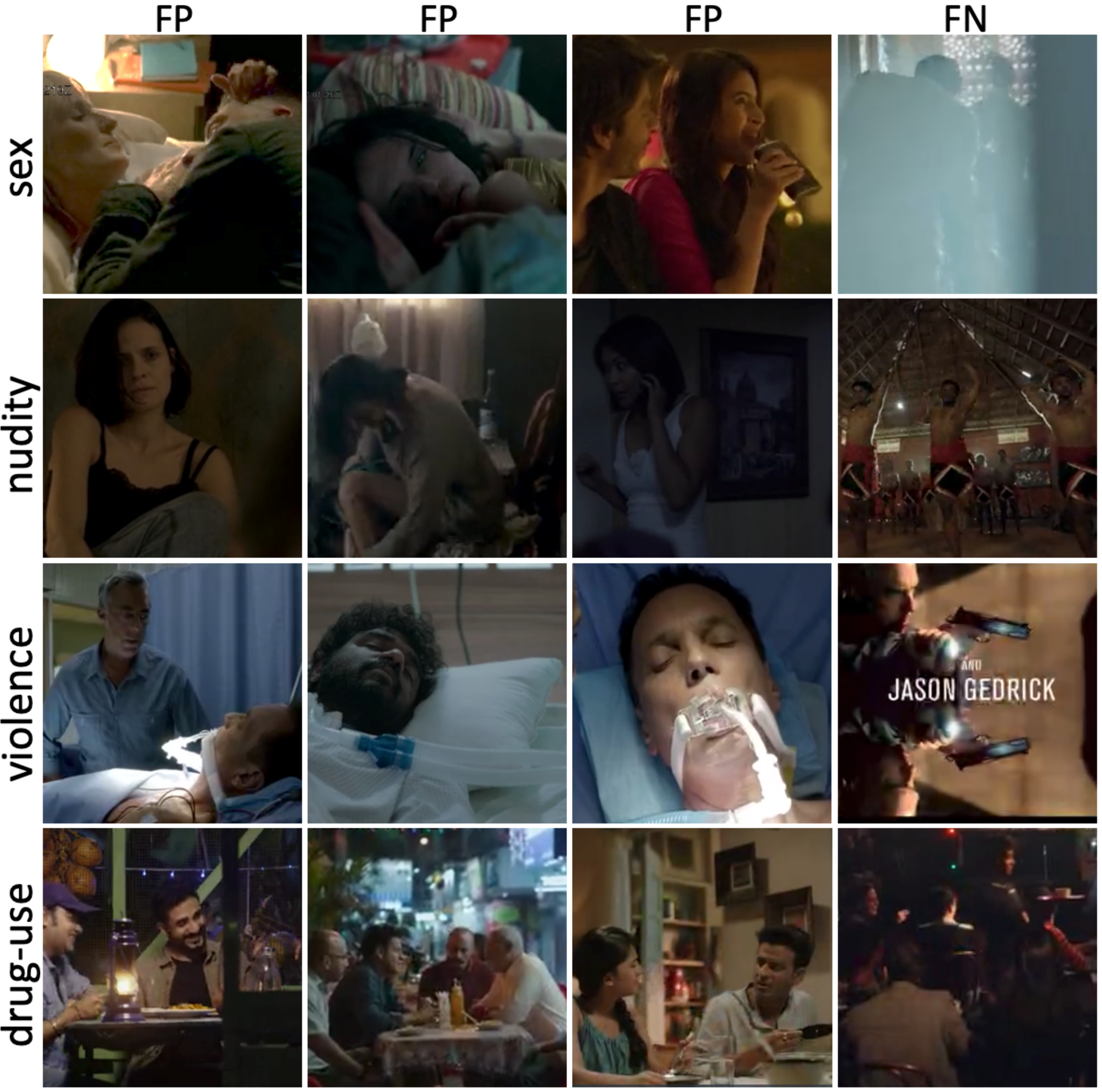}
    \caption{Representative False Positive (FP) and False Negative (FN) examples of model detections.}
    \label{fig:err}
      \vspace{-0.5cm}
\end{figure}

\noindent We present representative False Positive (FP) and False Negative (FN) examples of our model's output in Figure~\ref{fig:err}. As there are more background frames in our data-set, we show $3$ FP and $1$ FN examples for each activity-type to present a more balanced error-analysis of our model.



The FP examples of $\mathsf{sex}$ show that our model also managed to learn correlations between
sexual-activities and the presence of people especially when a bed is present in the background. The FN example for $\mathsf{sex}$ is mainly caused due to glass-reflection and the actors not facing the camera. The FP examples of $\mathsf{nudity}$ indicate that the model has difficulty understanding scenes when female actors appear with partially exposed body parts in poorly lit settings. The FN example for $\mathsf{nudity}$ is likely caused as the nude area of each person is relatively small and appears in a ceremonial environment; a setting which mostly does not contain nudity.





For action-type $\mathsf{violence}$, we notice that the model tends to treat frames in
emergency room as $\mathsf{violence}$ as shown in the first $3$ FP examples,
potentially caused by the correlation between injuries and medical first
aid. For false negatives, besides quick actions like a human holding a gun for
only two seconds in the intro of a video (as shown in the FN image for $\mathsf{violence}$), we also notice that fighting can be challenging to be
captured (such as when a man punches his fist on another person$’$s
face). Potential reason for this can be that optical flow based features are not used in
our model, which makes it hard for the model to learn the corresponding
motion. For $\mathsf{drug}$-$\mathsf{use}$, many FPs happen when a group of
people are sitting near the table but are not consuming alcohol, which may due to
the correlation between the existence of dining table and alcohol usage. Also,
the model is likely to predict positive when a human is holding a cigarette
shape object such as a chopstick, pencil or makeup pen which implies that
improvements can be achieved by introducing accurate object information. In
addition, the actions that the camera does not focus on are likely to be missed
by the model, such as the man in the top-left of the FN image for
$\mathsf{drug}$-$\mathsf{use}$ smoking in the background.

\vspace{-0.25cm}
\section{Conclusions}
\vspace{-0.25cm}
\noindent In this work, we investigated how to leverage video-level labels in conjunction with spare clip-level labels to train accurate temporal localization model for age-appropriate activities in long-form videos. We presented a novel approach that decomposes frame-level prediction into a binary classification module for identifying frames containing any type of sensitive activity, followed by predicting the action classes for each of the detected foreground-frames. We demonstrated the overall effectiveness of our approach using 4,1234 movies and TV episodes making it the largest-scale empirical analysis of age-appropriate activity localization in long-form videos ever published. We showed that our approach offers $107.2$\% relative mAP improvement (from $5.5$\% to $11.4$\%) over existing state-of-the-art activity-localization approaches. Going forward, we plan to experiment other types of visual features, such as optical flow, and explore other feature modalities, such as audio, in addition to visual signal to improve the detection accuracy.  


{\small
	\bibliographystyle{ieee_fullname}
	\bibliography{TALCompliance_arxiv2022}
}

\end{document}